\pdfoutput=1

\documentclass[11pt]{article}

\usepackage[final]{acl}

\usepackage{times}
\usepackage{latexsym}
\usepackage{alltt}

\usepackage[T1]{fontenc}

\usepackage[utf8]{inputenc}
\usepackage{amsmath}

\usepackage{microtype}

\usepackage{inconsolata}

\usepackage{graphicx}

\title{Towards Long Context Hallucination Detection}

\author{
 \textbf{Siyi Liu \thanks{Work done during internship at AWS AI Labs} \textsuperscript{1,2}},
 \textbf{Kishaloy Halder\textsuperscript{1}},
 \textbf{Zheng Qi\textsuperscript{1}},
\\
 \textbf{Wei Xiao\textsuperscript{1}},
 \textbf{Nikolaos Pappas\textsuperscript{1}},
 \textbf{Phu Mon Htut\textsuperscript{1}},
\\
 \textbf{Neha Anna John\textsuperscript{1}},
 \textbf{Yassine Benajiba\textsuperscript{1}},
 \textbf{Dan Roth\textsuperscript{1,2}}
\\
\\
 \textsuperscript{1}AWS AI Labs,
 \textsuperscript{2}University of Pennsylvania,
\\
 \texttt{
 siyiliu@seas.upenn.edu
 }
}

\begin{document}
\maketitle
\begin{abstract}

Large Language Models (LLMs) have demonstrated remarkable performance across various tasks. However, they are prone to \textit{contextual hallucination}, generating information that is either unsubstantiated or contradictory to the given context. Although many studies have investigated contextual hallucinations in LLMs, addressing them in long-context inputs remains an open problem. In this work, we take an initial step toward solving this problem by constructing a dataset specifically designed for long-context hallucination detection. Furthermore, we propose a novel architecture that enables pre-trained encoder models, such as BERT, to process long contexts and effectively detect contextual hallucinations through a decomposition and aggregation mechanism. Our experimental results show that the proposed architecture significantly outperforms previous models of similar size as well as LLM-based models across various metrics, while providing substantially faster inference. We publicly release our dataset and code to promote research along the same line\footnote{\url{https://github.com/amazon-science/long-context-hallucination-detection}}.

\end{abstract}

\section{Introduction}
Large language models (LLMs) have demonstrated potential in generative and knowledge-intensive tasks, such as question-answering (QA) and summarization. Despite these advancements, their practical deployment remains challenging, particularly due to the issue of \textit{hallucination}, wherein models generate content that appears plausible but is factually incorrect or unsubstantiated. In this work, we focus on a specific type of \textit{hallucination}—\textit{contextual hallucination}—in which models produce information that is either unsubstantiated or contradictory to the given context.

Previous work has studied contextual hallucination detection through the lens of Natural Language Inference (NLI): given a pair of input texts context and response, a generated response is considered faithful and free of hallucinations only when it is logically entailed by the context \cite{maynez2020faithfulnessfactualityabstractivesummarization, kryscinski-etal-2020-evaluating, 10.1162/tacl_a_00373,zha2023alignscoreevaluatingfactualconsistency}. Some studies explore contextual hallucination detection by training small, encoder models like BERT \cite{devlin2019bertpretrainingdeepbidirectional} or RoBERTa \cite{liu2019robertarobustlyoptimizedbert} on NLI datasets \cite{kryscinski-etal-2020-evaluating, zha2023alignscoreevaluatingfactualconsistency}; some other studies take a LLM-based approach and prompt LLMs to assess whether such hallucinations are present \cite{chang2024booookscoresystematicexplorationbooklength, hu2024refcheckerreferencebasedfinegrainedhallucination}. However, both lines of work encounter challenges when addressing longer contexts. For instance, BERT-based models for contextual hallucination detection are constrained by a maximum input length of 512 tokens, while LLM-based prompting for evaluating the faithfulness of responses to long contexts is not only expensive but also empirically suboptimal \cite{kim2024fablesevaluatingfaithfulnesscontent}.

In this work, we introduce a novel architecture that enables pre-trained encoder models, such as BERT, to process longer contexts and effectively detect contextual hallucinations through a decomposition and aggregation mechanism. Our model begins by decomposing the long input contexts and responses into smaller chunks. It then generates deep representations for each chunk using a backbone encoder model. Finally, it aggregates these chunk-level representations through a learned attention and pooling layer to create a holistic representation of both the context and response chunks to evaluate hallucination. Due to the scarcity of available datasets in long-context hallucination detection, we develop a prompting workflow that introduces contextual hallucinations into an existing long document summarization dataset, BookSum \cite{booksum}, to empirically evaluate our proposed architecture. Our experimental results demonstrate that the proposed architecture significantly outperforms prior approaches while offering substantially faster inference.


\section{Related Work}
LLMs are prone to hallucinations, wherein they generate content that appears plausible but lacks factual accuracy or is nonsensical. \citet{10.1145/3703155} have broadly categorized hallucinations into two primary types: \textit{factuality hallucination} and \textit{faithfulness hallucination}. Previous work on hallucinations primarily focused on \textit{factuality hallucination} \cite{min-etal-2023-factscore, chen2024complexclaimverificationevidence,zha2023alignscoreevaluatingfactualconsistency}, often leveraging the internal representations of models to detect such hallucinations \cite{azaria-mitchell-2023-internal, burns2024discoveringlatentknowledgelanguage}. On the other hand, \textit{faithfulness hallucination}, also known as \textit{contextual hallucination} \cite{chuang2024lookbacklensdetectingmitigating}, remains relatively understudied. In this paper, we specifically focus on this type of hallucination, where the model generates information that is either unsubstantiated or contradictory to the given context, without assessing the factual accuracy of generated content. 

Previous work on \textit{contextual hallucination} detection and mitigation often relies on access to a model’s internal states, such as attention maps \cite{chuang2024lookbacklensdetectingmitigating} or output distributions \cite{shi2023trustingevidencehallucinatecontextaware}. In this study, we focus on black-box models, where internal states are inaccessible, requiring hallucination detection to be performed solely based on the input context and generated response. As a result, methods that depend on internal model states are not applicable.
Another approach to hallucination detection leverages Natural Language Inference (NLI): a generated response is considered faithful and free of hallucinations only if it is logically entailed by the provided context \cite{maynez2020faithfulnessfactualityabstractivesummarization, kryscinski-etal-2020-evaluating, 10.1162/tacl_a_00373, zha2023alignscoreevaluatingfactualconsistency}. However, existing NLI-based methods predominantly rely on small encoder-only models such as BERT \cite{devlin2019bertpretrainingdeepbidirectional} or RoBERTa \cite{liu2019robertarobustlyoptimizedbert}, which are constrained by a maximum input length of 512 tokens, limiting their applicability to long-context scenarios.
A separate line of work explores the use of Large Language Models (LLMs) for hallucination assessment \cite{chang2024booookscoresystematicexplorationbooklength, hu2024refcheckerreferencebasedfinegrainedhallucination}. While these methods can provide strong performance, they are computationally expensive and slow, making their deployment in practical settings challenging. 

Research on long-form hallucination detection remains limited. \citet{song2024veriscoreevaluatingfactualityverifiable} propose a metric for assessing the factuality of verifiable claims in long-form text generation, while \citet{kim2024fablesevaluatingfaithfulnesscontent} conduct a large-scale human evaluation of faithfulness and content selection in LLM-generated summaries of fictional books. Our work bridges this gap by introducing a new dataset for both training and evaluation, along with a novel architecture that extends the capability of pre-trained encoder models, such as BERT, to process longer contexts and effectively detect contextual hallucinations with improved efficiency.


\section{Problem Definition}
\label{sec:definition}

In this work, we investigate the problem of long-context hallucination detection. Our objective is to develop a model that can effectively and efficiently detect \textit{contextual hallucinations} given solely a pair of input texts: a context and a corresponding response, without access to the internal states of models. Specifically, we focus on cases where the context is long-form, which presents additional challenges for models in terms of processing and making inferences within a short time frame.

We define the \textit{contextual hallucinations} examined in this study as follows: given a document, a response is considered to contain hallucinations if and only if (a) it introduces unsubstantiated information that is not grounded in the context, or (b) it presents information that contradicts the context. The models are expected to perform a binary classification to determine whether the response hallucinates relative to the context, regardless of the specific type of hallucination.

To empirically evaluate our models within this problem setting, we conduct experiments on the task of long-document summarization, where the context consists of a long document about a book and the response is a corresponding summary. However, we posit that our hallucination injection framework and model design can also generalize to other domains involving long-context hallucination detection such as dialogue systems.

\section{Dataset Collection}

We consider the task of book summarization to support our experiments and construct our dataset from BookSum \cite{booksum}. This dataset includes varying levels of document-summary pairs, including book-level, chapter-level, and paragraph-level pairs. In our study, we focus on chapter-level document-summary pairs, as they align more closely with our research interests. Chapter-level documents have on average 5,101 tokens, and summaries have on average 505 tokens. The dataset only provides expert written, ground-truth summaries for the different levels of documents. We synthesize a hallucinatory subset by injecting some hallucination for certain pairs in the dataset. To create a balanced dataset, we introduce hallucinations with a 50\% probability while iterating through the dataset. Each time we introduce a hallucination, we randomly select one type of hallucination from the two categories introduced in Section \ref{sec:injection}. The statistics of our dataset is shown in Table \ref{tab:data_stats}.

\subsection{Hallucination Injection}
\label{sec:injection}
We develop a prompting workflow that supports us to introduce hallucination to our dataset of long document summarization. We consider two following types of hallucination as introduced in Section \ref{sec:definition}. The exact prompts we use for this process are shown in Appendix \ref{appendix:prompt}.

\paragraph{Baseless Information Hallucination} We prompt GPT-4o to 
\textit{"add a complete sentence that is related to the topic but introduces some new information you make up ..."}.

\paragraph{Contradictory Information Hallucination}
We prompt GPT-4o to \textit{"rewrite one sentence completely so that it utterly contradicts from its original sentence ..."}.


\begin{table}[]
\begin{tabular} {ccc}
\hline
Split  & \# of Examples& \% of hallucinations  \\
\hline
Train & 5,653 & 51\%  \\

Dev & 854   & 48\%\\

Test & 950 & 52\%\\

\hline
\end{tabular}
\centering
\caption{The statistics of our constructed dataset.}
\label{tab:data_stats}
\end{table}

\subsection{Dataset Verification}

Given that our dataset consists of extensive book chapters, manual verification is both labor-intensive and costly, as it requires a thorough reading of the content. Therefore, we utilize Perplexity score as an automated estimate to assess the coherence and fluency of summaries following the introduction of hallucinations. Perplexity is defined as the exponentiated average negative log-likelihood of a sequence and is popularly used as a measure to evaluate the performance of a language model as well as the quality of generations. It quantifies how well a probabilistic model predicts a sequence of words. A lower perplexity score indicates that the language model assesses the sequence of text as being more aligned with its predicted probabilities, reflecting better coherence and fluency. We calculate the perplexity score of a summary as follows:
$\text{Perplexity} = \exp\left(-\frac{1}{N} \sum_{i=1}^{N} \log P(w_i) \right)$.

We utilize Llama-3.2-1B to compute the average perplexity scores for both the original summaries and the summaries after the introduction of hallucination. Interestingly, we observe that the average perplexity score decreases from 18.52 to 18.26 after the injection of hallucinations, indicating a high quality of our data augmentation process.

\begin{figure}
    \centering
    \includegraphics[width=1\linewidth]{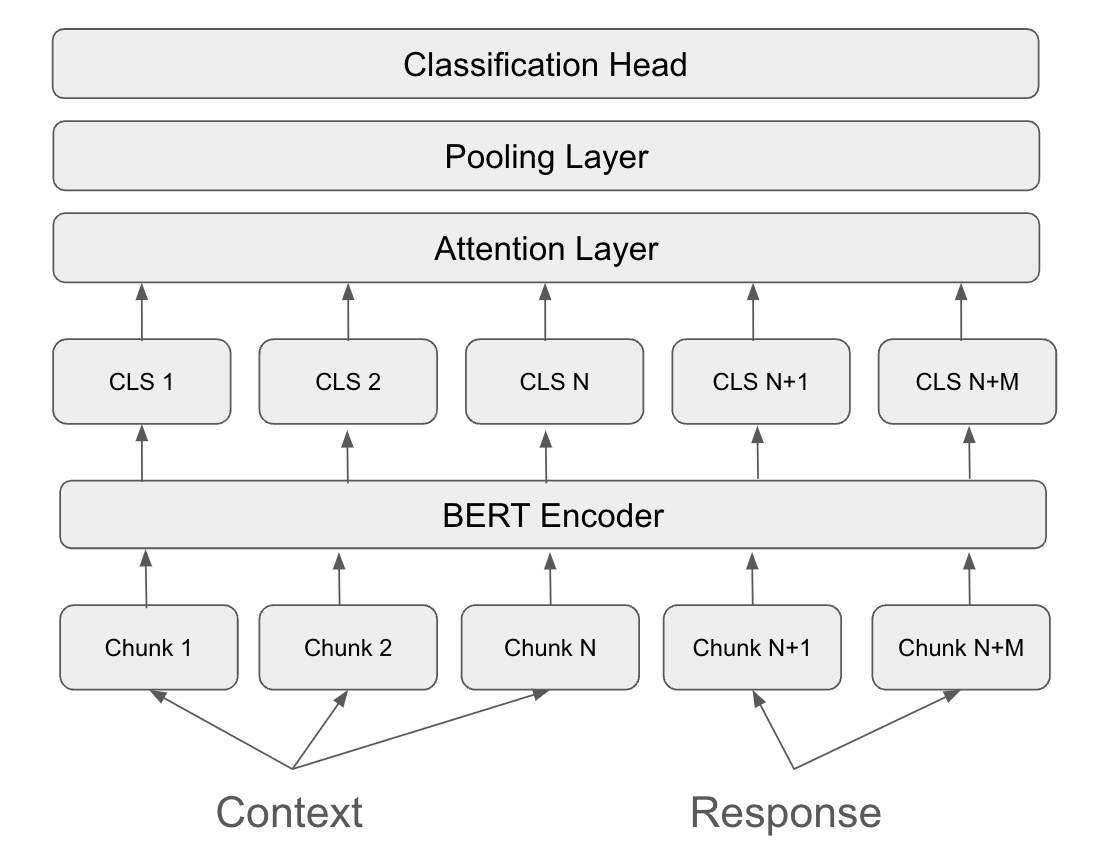}
    \caption{The structure of our proposed architecture. In the attention layer, we add a new token of [CLS] at the beginning of all chunk-level CLS representations to be used as a pooled representation for the whole input, and a [SEP] between the context chunk representations and the response chunk representations to distinguish them.}
    \label{fig:model}
\end{figure}

\section{Our Method}
\label{sec:method}
The primary obstacle preventing BERT-based models from effectively processing long documents is the computation of the full quadratic attention matrix, which incurs \(O(n^2)\) time and memory complexity, where $n$ represents the input sequence length. Intuitively, each token must attend to all other tokens to develop robust representations of the input texts. To tackle this challenge, we propose an architecture that employs a decomposition and aggregation strategy. 
The structure of our model is shown in Figure \ref{fig:model}. Given a pair of input texts—context and response—we first decompose them into fixed length chunks for both the context and response. Each chunk is then processed through a pre-trained BERT encoder to obtain their corresponding CLS representations. Subsequently, we employ an attention layer to learn which chunks are most prominent for assessing the presence of hallucinations in the response with respect to the context. Finally, we utilize a pooling layer to obtain a holistic representation of all chunks for the purpose of classification. We provide further experimental details regarding chunk sizes, the number of chunks, and various other hyperparameters and architectural design choices in Section \ref{sec:experiment} and Appendix \ref{appendix:details}.


Our proposed architecture offers several advantages: 1. Our framework does not necessitate any further pretraining and can be implemented on top of existing encoder models. In contrast, previous approaches for long-context processing, such as Hierarchical Attention Transformer (HAT) \cite{chalkidis2022explorationhierarchicalattentiontransformers} or Longformer \cite{beltagy2020longformerlongdocumenttransformer} require pretraining on long-form texts, which can be computationally expensive. Our model circumvents this requirement, enabling the use of any encoder model as the backbone for fine-tuning on domain-specific tasks, such as long-context hallucination detection. 2. Theoretically, our model can accommodate very long contexts by continually adding layers of decomposition and aggregation (one layer can process up to $512$ chunks $\times 512$ chunk size of tokens). Given a fixed chunk length $c$ (e.g. 512), the computation complexity of our model is  $O(k^2)$, where $k$ denotes the number of chunks and $k=\frac{n}{c}$. This represents a significant improvement over the \(O(n^2)\) complexity of BERT.



\section{Experiment}
\label{sec:experiment}
We conduct experiments using our constructed dataset and compare the performance of our proposed model with that of previous approaches.

\subsection{Models}
\paragraph{Longformer}
Longformer is a modified Transformer architecture with a self-attention operation that scales linearly with the sequence length, making it versatile for processing long documents \cite{beltagy2020longformerlongdocumenttransformer}. We finetune a pre-trained Longformer model using our dataset for model comparison.

\paragraph{Hierarchical Attention Transformer (HAT)}
Hierarchical Attention Transformers (HATs) employ a multilevel attention mechanism consists of segment-wise attention followed by cross-segment attention to effectively handle long documents \cite{chalkidis2022explorationhierarchicalattentiontransformers}. We finetune a pre-trained HAT model using our dataset for our experiments.

\paragraph{Alignscore}
Alignscore is a RoBERTa model trained on a general function that assesses the information alignment between two arbitrary text pieces. Its training incorporates a wide range of data sources, resulting in 4.7 million training examples derived from seven well-established tasks: Natural Language Inference (NLI), Question Answering (QA), paraphrasing, fact verification, information retrieval, semantic similarity, and summarization.
\cite{zha2023alignscoreevaluatingfactualconsistency}. 
The model can infer with arbitrarily long texts; however, it cannot be trained on texts longer than 512 tokens. The authors also present it as an off-the-shelf metric, given that it has been trained on a substantial amount of factual consistency data. Therefore, we evaluate the model off-the-shelf without any additional training in this study.

\begin{figure}
    \centering
    \includegraphics[width=1\linewidth]{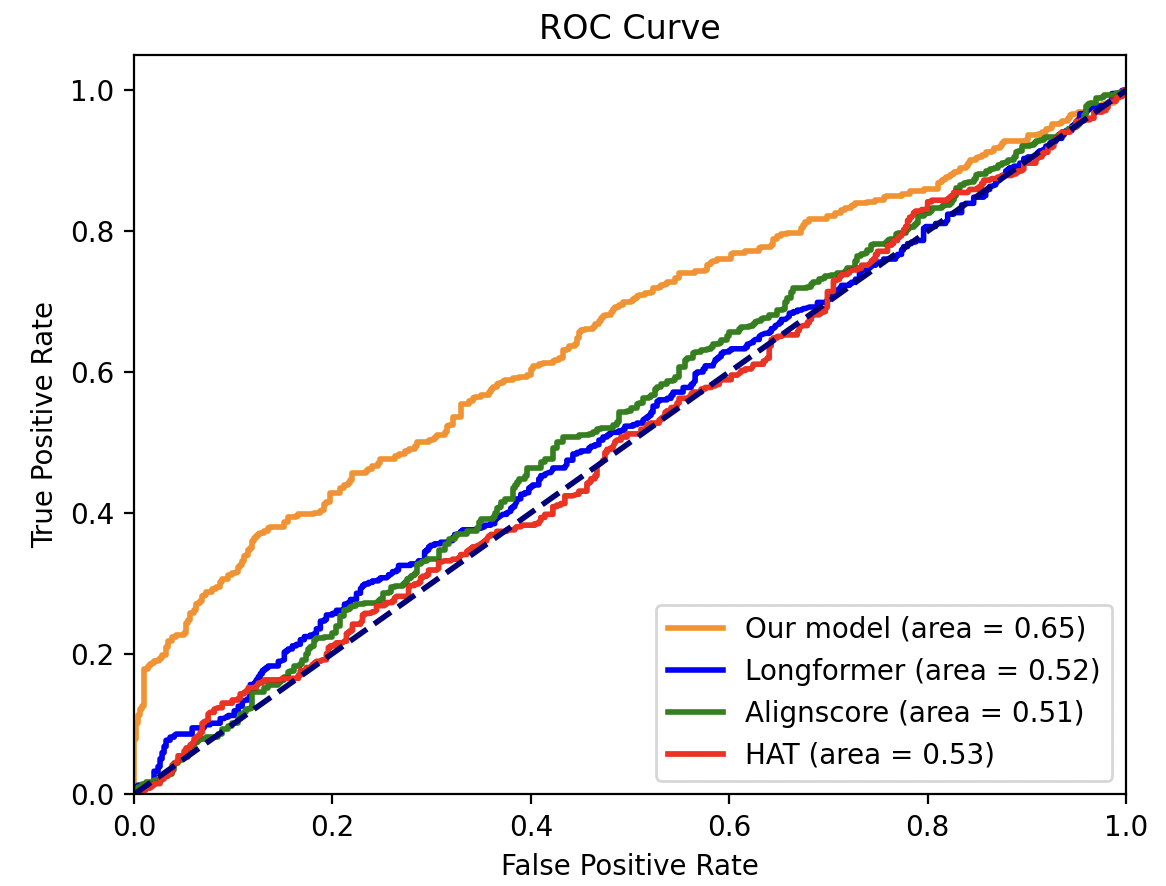}
    \caption{ROC AUC results of Hierarchical Attention Transformer (HAT), Longformer, Alignscore, and our method.}
    \label{fig:rocauc}
\end{figure}

\paragraph{RefChecker}
RefChecker introduces
claim-triplets to represent claims in LLM responses, aiming to detect fine-grained hallucinations \cite{hu2024refcheckerreferencebasedfinegrainedhallucination}. This framework first prompts an LLM to extract claims from the response, and then prompt an LLM another time to compare each of the claim to the context to predict hallucination. We use GPT-4o-mini as the LLM backbone for both the extractor and checker in their framework.

\paragraph{GPT-4o}
We zero-shot prompt GPT-4o-mini with specific instructions and definitions of our task to predict hallucinations as a strong baseline. The exact prompt that we use is shown in Appendix \ref{appendix:prompt}.

\paragraph{Our Model}
The structure of our model is described in Section \ref{sec:method}. 
More experimental details about our model are discussed in Appendix \ref{appendix:details}.

\subsection{Results}
We present the Receiver Operating Characteristic (ROC) Curve and the ROC Area Under the Curve (AUC) score in Figure \ref{fig:rocauc}. Due to the black-box nature of LLM-based models, we are unable to obtain their predicted scores, so only the results from encoder models are displayed. We see that all baseline models lack discriminative ability in terms of detecting hallucination with long context: state-of-the-art metrics in factual consistency evaluation like AlignScore fail to adapt to long-form texts; Longformer and HAT also exhibit insufficient expressive capacity to distinguish hallucinations, despite being pre-trained on long-form texts and then finetuned on the same training set as our model utill converged. In contrast, our model demonstrates strong performance on this task, without any pre-training on long-form or factual consistency data.

\begin{table}[]

\begin{tabular}{c@{\hskip 4pt}|@{\hskip 3pt}ccc}
\hline
Model  & \small \textsc{Precision} & \small \textsc{Recall} & \small \textsc{Latency}  \\

\hline
HAT  & 48.42 & 70.55 & \textbf{41.01} \\
Longformer & 47.89 & \textbf{87.47} & 18.15 \\
Alignscore  & 50.09 & 60.00 & 1.44\\
Refchecker & 52.13 & 51.21  & 0.15\\ 
GPT-4o & \underline{53.11} & \underline{78.68} & 0.79\\
\hline
Our Model& \textbf{54.50} & 73.19 & \underline{18.62}\\

\hline
\end{tabular}
\centering
\caption{Results of all of the models we tested. Latency is computed as the number of samples processed per second at inference time, the higher the faster. The \textbf{bolded} numbers represent the best performance across all models and the \underline{underlined} numbers represent the second best. See more details about hyperparameter choices as well as how latency is computed in Appendix \ref{appendix:details}.}
\label{tab:results}
\end{table}

We show the precision, recall score and inference latency of our model and all baseline models in Table \ref{tab:results}. Notably, Longformer exhibits high recall but low precision, indicating that it tends to overpredict the positive class, leading to a high number of false positives. Additionally, while Refchecker takes considerably more time for inference by extracting and verifying individual claims, it performs worse than GPT-4o, despite using the same backbone LLM. This suggests that traditional approaches to hallucination detection, which rely on splitting inputs into claims and verifying each claim to produce an aggregated score, may not be as effective when applied to long-context inputs. This observation aligns with the suboptimal performance of AlignScore on our dataset, as its approach mirrors this method. Our model, on the other hand, matches GPT-4o in precision and recall but achieves 20x faster inference times, making it more applicable for real-world deployment. More details of how we measure the inference latency are discussed in Appendix \ref{appendix:details}.

In addition, we incorporate Balanced Accuracy and Matthews Correlation Coefficient (MCC) to gain a deeper understanding of the performance gap between our proposed approach and LLM-based methods, as these aspects were not clearly reflected in our previous evaluations. Balanced Accuracy is a widely used metric in classification tasks \cite{5597285}, calculated as the average of a model’s sensitivity and specificity. MCC is a robust statistical measure that yields a high score only when predictions achieve strong performance across all four categories of the confusion matrix. It produces values in the range of $[-1,1]$ and has been shown to provide a more informative and truthful score in evaluating binary classifications than accuracy and F1 score \cite{chicco2020advantages}. As presented in Table \ref{tab:additional_metrics}, our proposed method substantially outperforms LLM-based approaches on our dataset, highlighting both its superior performance and its efficiency in inference speed.

\begin{table}[]

\begin{tabular} {c|ccc}
\hline
Model  & \small \textsc{Balanced Accuracy} & \small \textsc{MCC} \\

\hline
Refchecker & 53.99 & 0.08  \\ 
GPT-4o & 57.42 & 0.16 \\
\hline
Our Model& \textbf{67.22} & \textbf{0.26} \\

\hline
\end{tabular}
\centering
\caption{Results of LLM-based baselines and our model on balanced accuracy and matthews correlation coefficient.}
\label{tab:additional_metrics}
\end{table}





\section{Conclusion}

We introduce a new dataset and propose a novel architecture to investigate long-context hallucination detection. Our experimental results demonstrate that the proposed method significantly outperforms previous approaches across various metrics while achieving substantially faster inference. We anticipate that our dataset and research artifacts will serve as valuable resources for future studies in this domain.

\paragraph{Limitations}
One limitation of our work is that our proposed model requires in-domain training for a specific domain. This is different from prompting with LLMs. However, our proposed prompting workflow of hallucination injection makes it easy to obtain high-quality training data for other domains (e.g. dialogue) as well to support the training of our model in these areas, and then our model will have faster inference time in deployment with on par performance with strong LLMs. 

Although theoretically our model can accommodate extremely long contexts by continuously adding layers of decomposition and aggregation—where each layer can process up to $512$ chunks $\times 512$ tokens per chunk—we have not empirically tested its performance on datasets containing texts longer than those used in our study. The extent to which the observed performance improvements generalize to even longer texts, as well as the model’s applicability to different domains and genres, remains an open question.

\paragraph{Ethics Statement}

In this work, we utilize publicly available datasets in accordance with their intended usage and licenses. We propose a method for detecting contextual hallucinations in long-context scenarios, which has the potential to positively impact downstream applications such as question answering and summarization by identifying hallucinated content. However, this remains a highly challenging problem, and our model may not always produce accurate results. Therefore, careful consideration and thorough evaluation are necessary before deploying it, particularly in sensitive domains. Additionally, although minimal, our constructed dataset includes machine-generated data, which may introduce potential biases and associated risks. Careful data evaluation and verification are necessary to ensure the dataset's reliability and fairness, particularly in mitigating any unintended effects of machine-generated information.

\section*{Acknowledgments}
This work was conducted during Siyi Liu’s internship at Amazon Web Services (AWS) and was fully funded by AWS. We sincerely appreciate the valuable feedback provided by the AWS team and extend our gratitude to the ARR reviewers and editors for their insightful and constructive comments.

\bibliography{custom}

\appendix
\section{Experiment Details}
\label{appendix:details}
\subsection{Training Details}
We train our model with the Huggingface Transformers and Accelerate package. We use Amazon Elastic Compute Cloud (Amazon EC2) for our training experiments. We use one p4d.24xlarge instance for the training. It has 8 NVIDIA A100 GPUs with 40.0 GB GPU memory each. The optimal hyperparamters we find for our model is 40 chunks in total, 32 for context and 8 for response, each with a chunk size of 256. We train our model with 2e-6 learning rate, 0.1 weight decay, 1000 warm up steps, and 100 epochs. We train with only the first 1,000 examples for our model as it already shows good performance in the validation set. We use pre-trained Roberta-large as our backbone encoder model and a randomly initialized Roberta Attention layer. All parameters in the architecture are being optimized. In the attention layer, we add a new token of [CLS] at the beginning of all chunk-level CLS representations to be used as a pooled representation for the whole input, and a [SEP] between the context chunk representations and the response chunk representations to distinguish them.

\subsection{Inference Latency}
HAT, Longformer, and our model inference with 8 GPUs (data parallel) with a batch size of 4. However, the codebase provided by the authors of Alignscore doesn’t support multi-gpu inference with longer texts and also doesn’t support batching. So the inference latency of AlignScore is computed as their inference time with one gpu and batch size of one multiplied by 32 as an estimate. Inference time of GPT-4o and Refchecker depends on API calls to OpenAI and may differ from time to time due to network, API availability, and some other reasons.

\section{Dataset Examples}
The whole chapter is too long to present, so here we show examples of original summary from the BookSum dataset, as well as summary after our hallucination injection. We highlight the specific sentence that was rewritten or added in different colors.

\paragraph{Original Summary}
Any state--old, new, whatever--needs good laws and good armed forces. Since you can't have good armed forces without good law, let's just say you need a good army. There are four types of armies you could have: a local army, mercenaries, auxiliaries , or some kind of mixture. \textcolor{blue}{First things first: mercenaries and auxiliary armies are useless. Just don't do it.} Mercenaries are only interested in the money and are not reliable. That's how Italy got into trouble--occupation by France and Spain--in the first place. Plus, if a mercenary leader is good then you have to be afraid that he will turn against you, and if he is bad he will make you lose anyway. So, no good. Good armies? Citizen armies. Look at Rome. Look at Sparta! Carthage used mercenaries, and guess what happened to them? They got owned by Philip of Macedonia, Alexander the Great's dad. Don't be like Carthage. Machiavelli gives us a bunch of examples, but the basic takeaway here is mercenaries = bad. They are lazy. They are expensive. They kill during wartime. And they don't even defend their camps. Machiavelli has a little bit of an axe to grind about this problem, since he has wanted Italy to stop relying on mercenaries forever, but no one would listen to him.

\paragraph{Summary After Injection of Contradictory Information}
Any state--old, new, whatever--needs good laws and good armed forces. Since you can't have good armed forces without good law, let's just say you need a good army. There are four types of armies you could have: a local army, mercenaries, auxiliaries, or some kind of mixture. \textcolor{red}{First things first: mercenaries and auxiliary armies are essential. Always rely on them.} Mercenaries are only interested in the money and are not reliable. That's how Italy got into trouble--occupation by France and Spain--in the first place. Plus, if a mercenary leader is good then you have to be afraid that he will turn against you, and if he is bad he will make you lose anyway. So, no good. Good armies? Citizen armies. Look at Rome. Look at Sparta! Carthage used mercenaries, and guess what happened to them? They got owned by Philip of Macedonia, Alexander the Great's dad. Don't be like Carthage. Machiavelli gives us a bunch of examples, but the basic takeaway here is mercenaries = bad. They are lazy. They are expensive. They kill during wartime. And they don't even defend their camps. Machiavelli has a little bit of an axe to grind about this problem, since he has wanted Italy to stop relying on mercenaries forever, but no one would listen to him.
\paragraph{Original Summary}
Chapter X is entitled "How to Measure the Strength of Any Prince's State. Here Machiavelli adopts a decidedly militaristic tone. Princes, he writes, are better off when they can assemble an army and stand up against attackers; once again, Cesare Borgia is cited as a perfect example. Machiavelli addresses the majority of this chapter to the other class of princes: "those who can't take the field against their foes, but have to hide behind their walls and defend themselves there. What should these more vulnerable princes do. They should keep their cities well-fortified; they should ignore the rural areas and focus their defense efforts on the urban centers; and they should be careful not to earn the people's hatred. A prudent prince is able to keep his subjects loyal to him and in good spirits during a siege. The burden during a siege is often on the besieger; he can almost never afford to wage a siege and do nothing else for a year. Defense, therefore, can consist of slowing the attacker down, wearing him out. Machiavelli cites the cities in Germany as examples of good fortification. These cities have moats, walls, artillery, public warehouses of food, drink, and fuel, and large supplies of raw materials in reserve to keep workers busy and economies going during a siege


\paragraph{Summary After Injection of Baseless New Information}

Chapter X is entitled "How to Measure the Strength of Any Prince's State." Here Machiavelli adopts a decidedly militaristic tone. Princes, he writes, are better off when they can assemble an army and stand up against attackers; once again, Cesare Borgia is cited as a perfect example. Machiavelli addresses the majority of this chapter to the other class of princes: "those who can't take the field against their foes, but have to hide behind their walls and defend themselves there." What should these more vulnerable princes do? They should keep their cities well-fortified; they should ignore the rural areas and focus their defense efforts on the urban centers; and they should be careful not to earn the people's hatred. \textcolor{red}{He notes that a well-designed urban area can serve as a formidable defense mechanism, with strategically placed fortifications and supply depots.} A prudent prince is able to keep his subjects loyal to him and in good spirits during a siege. The burden during a siege is often on the besieger; he can almost never afford to wage a siege and do nothing else for a year. Defense, therefore, can consist of slowing the attacker down, wearing him out. Machiavelli cites the cities in Germany as examples of good fortification. These cities have moats, walls, artillery, public warehouses of food, drink, and fuel, and large supplies of raw materials in reserve to keep workers busy and economies going during a siege.

\label{appendix:example}

\section{GPT-4o Prompts}
\label{appendix:prompt}

\paragraph{Prompts Used to Introduce Baseless Information Hallucination}
"Add a complete sentence that is related to the topic but introduces some new information you make up. You can add the sentence anywhere in the paragraph but make sure it is a complete sentence and the paragraph is coherent. Reply with the whole paragraph that includes the sentence you added."

\paragraph{Prompts Used to Introduce Contradictory Information Hallucination}

"Given the paragraph, rewrite one sentence completely so that it utterly contradicts from its original sentence. You can choose any sentence in the paragraph but make sure the paragraph is still coherent and now has a claim that contradicts the original paragraph. Reply with the whole paragraph after the change."

\paragraph{Prompts Used to Run GPT-4o-mini Experiments}

"You will be given a document and a summary. Your task is to determine whether the summary is faithful or unfaithful to the information provided in the document. If the summary contains any statements that contradict the information given in the document, or if it includes information not present or implied by the document, reply 'unfaithful'. Otherwise, reply 'faithful'."
\label{appendix:example}



\end{document}